\documentclass[sigconf]{acmart}

\AtBeginDocument{%
  }

\usepackage[inline]{enumitem}
\usepackage{algorithmic}
\usepackage{graphicx}
\usepackage{textcomp}
\usepackage{xcolor}

\usepackage{witharrows}

\usepackage[flushleft]{threeparttable}
\usepackage{tablefootnote}
\usepackage{array}
\newcolumntype{P}[1]{>{\centering\arraybackslash}p{#1}}
\usepackage{multicol}
\usepackage{multirow}
\usepackage{tabulary}
\usepackage{colortbl}
\definecolor{mygray}{gray}{0.90}
\usepackage{makecell}
\usepackage{booktabs}
\usepackage{subcaption}

\usepackage{stmaryrd}

\usepackage{float}
\usepackage{pifont}

\usepackage{listings}
\usepackage{xcolor}

\usepackage{arydshln}
\usepackage[switch,columnwise]{lineno}

\copyrightyear{2025}
\acmYear{2025}
\setcopyright{cc}
\setcctype{by}\setcctype{by}
\acmConference[ICMI Companion '25]{Companion Proceedings of the 27th
International Conference on Multimodal Interaction}{October 13--17,
2025}{Canberra, ACT, Australia}
\acmBooktitle{Companion Proceedings of the 27th International Conference on
Multimodal Interaction (ICMI Companion '25), October 13--17, 2025, Canberra,
ACT, Australia}
\acmDOI{10.1145/3747327.3764782}
\acmISBN{979-8-4007-2076-5/2025/10}

\begin{document}

\title{Efficient Pain Recognition via Respiration Signals: A Single Cross-Attention Transformer Multi-Window Fusion Pipeline}

\author{Stefanos Gkikas}
\email{gkikas@ics.forth.gr}
\orcid{0000-0002-4123-1302}
\affiliation{%
  \institution{Foundation for Research \& Technology-Hellas}
  \city{Heraklion}
  \country{Greece}
}

\author{Ioannis Kyprakis }
\email{ikyprakis@ics.forth.gr}
\orcid{0009-0000-0711-5108}
\affiliation{%
  \institution{Foundation for Research \& Technology-Hellas}
  \city{Heraklion}
  \country{Greece}
}

%

\author{Manolis Tsiknakis}
\email{tsiknaki@ics.forth.gr}
\orcid{0000-0001-8454-1450}
\affiliation{%
  \institution{Foundation for Research \& Technology-Hellas and Hellenic Mediterranean University}
  \city{Heraklion}
  \country{Greece}
}


\begin{abstract}
Pain is a complex condition affecting a large portion of the population. Accurate and consistent evaluation is essential for individuals experiencing pain, and it supports the development of effective and advanced management strategies. 
Automatic pain assessment systems provide continuous monitoring and support clinical decision-making, aiming to reduce distress and prevent functional decline.
This study has been submitted to the \textit{Second Multimodal Sensing Grand Challenge for Next-Gen Pain Assessment (AI4PAIN)}. 
The proposed method introduces a pipeline that leverages respiration as the input signal and incorporates a highly efficient cross-attention transformer alongside a multi-windowing strategy.
Extensive experiments demonstrate that respiration is a valuable physiological modality for pain assessment. Moreover, experiments revealed that compact and efficient models, when properly optimized, can achieve strong performance, often surpassing larger counterparts.
The proposed multi-window approach effectively captures both short-term and long-term features, as well as global characteristics, thereby enhancing the model's representational capacity.

\end{abstract}




\maketitle

\section{Introduction}
Pain is a key evolutionary adaptation that signals possible injury or disease, playing a crucial role in safeguarding the organism’s physiological stability \cite{santiago_2022}.
Pain has been described as a \textit{\textquotedblleft Silent Public Health Epidemic\textquotedblright} \cite{katzman_gallagher_2024}, a term that reflects its widespread and often under-recognized impact. In the U.S., an estimated $50$ million people suffer daily from acute, chronic, or end-of-life pain, making it the leading reason for emergency room visits and medical consultations \cite{hhs_pain_2019}. Similar patterns are observed in Europe, where chronic pain leads to direct healthcare and indirect socioeconomic costs amounting to up to $10\%$ of the gross domestic product \cite{breivik_eisenberg_2013}.

Managing and assessing pain in patients with--or at risk of--medical instability presents significant clinical challenges, particularly when communication barriers are present \cite{puntilo_staannard_2022}.
Pain assessment strategies span a broad spectrum. Self-reporting methods, including numerical rating scales and questionnaires, remain the gold standard for assessing patient experiences. In parallel, behavioral indicators---such as facial expressions, vocalizations, and body movements-are also used to infer pain, particularly in non-communicative patients \cite{rojas_brown_2023}. Physiological measures, such as electrocardiography and skin conductance, further enhance assessment by providing objective insights into the body's response to pain \cite{gkikas_tsiknakis_slr_2023}.
A well-established bidirectional relationship exists between pain and respiration. Pain often triggers distinct respiratory responses---for example, an inspiratory gasp followed by breath-holding in reaction to sudden, acute pain; a sigh of relief upon pain relief; or episodes of hyperventilation during persistent, intense discomfort \cite{finesinger_mazick_1940}.
Despite these observations, the interaction between pain and respiration remains a complex phenomenon, posing significant challenges for research.
While acute pain is known to increase respiratory rate, flow, and volume, the effects of chronic pain on respiratory patterns are still not fully understood, and further investigation is needed 
\cite{jafari_courtois_2017}.

This study investigates the use of respiration as a standalone modality in an automatic pain assessment pipeline, aiming to explore its potential value---particularly from an engineering and machine learning perspective, where it has been largely unexplored. The proposed pipeline introduces an efficient single cross-attention transformer combined with a multi-window fusion approach designed to capture both local and global temporal features from the respiration signal.

\section{Related Work}
Over the past $15$ years, interest in automatic pain assessment has steadily increased, with developments progressing from classical image and signal processing techniques to more advanced deep learning-based approaches \cite{gkikas_phd_thesis_2025}. The majority of existing methods are video-based, aiming to capture behavioral cues through facial expressions, body movements, or other visual indicators and employing a wide range of modeling strategies \cite{gkikas_tsiknakis_embc, bargshady_hirachan_2024,gkikas_tsiknakis_thermal_2024,huang_dong_2022}.
While video-based approaches dominate the field, a considerable number of studies have also focused on biosignal-based methods, although to a lesser extent. These works have investigated the utility of various physiological signals, such as electrocardiography (ECG) \cite{gkikas_chatzaki_2022, gkikas_chatzaki_2023}, electromyography (EMG) \cite{pavlidou_tsiknakis_2025,patil_patil_2024,thiam_bellmann_kestler_2019,werner_hamadi_niese_2014}, electrodermal activity (EDA) \cite{aziz_joseph_2025,li_luo_2024,lu_ozek_kamarthi_2023,phan_iyortsuun_2023,ji_zhao_li_2023}, and brain activity through functional near-infrared spectroscopy (fNIRS) \cite{rojas_huang_2016, rojas_liao_2019,rojas_romero_2021,rojas_joseph_bargshady_2024,khan_sousani_2024,bargshady_aziz_2025}. For a more comprehensive analysis of biosignal modalities within automatic pain recognition frameworks, the reader is referred to \cite{khan_umar_2025}.
In addition, multimodal approaches combining behavioral and physiological data have gained increasing attention in recent years, with several studies demonstrating the benefits of integrating multiple sources of information to improve performance \cite{farmani_giuseppi_2025,gkikas_rojas_painformer_2025,zhi_yu_2019,jiang_rosio_2024,jiang_li_he_2024,
gkikas_tachos_2024,farmani_bargshady_2025}.

Studies incorporating respiration rate as an input signal are scarce.
The authors in \cite{rojas_hirachan_2023} extracted $30$ handcrafted and statistical features--including respiratory-rate-variability and respiratory-sinus-arrhythmia indices, amplitude metrics, and rate changes--and tested multiple classifiers. Respiration showed promise yet underperformed compared with electrodermal activity (EDA) and photoplethysmography (PPG).
In the study by Lin \textit{et al.} \cite{lin_xiao_2022}, the authors evaluated several modalities, including blood volume pulse (BVP), EMG, EEG, respiration, and others. BVP and EEG successfully distinguished pain states, and multimodal fusion appeared promising, but respiration alone did not achieve statistically significant separation. 
A subsequent study \cite{zhu_lin_2025} again highlighted BVP as the most influential sensor; respiration displayed stimulus-specific sensitivity yet had an inconsistent overall impact.
Winslow \textit{et al.} \cite{winslow_kwasinski_2022}, after computing respiratory and heart-rate-variability features and training logistic-regression classifiers, observed that respiration-rate changes were detectable but weaker pain discriminators than heart-rate-variability measures.
Similarly, Badura \textit{et al.} \cite{badura_2021} reported no significant contribution from respiration, whereas EDA most reliably reflected pain.
In contrast to previous studies, Cao \textit{et al.}\cite{cao_aqajari_2021} utilized respiratory rate derived from wristband-recorded PPG signals in postoperative patients and achieved strong pain-detection performance---perhaps due to the binary nature of their classification task.
Similarly, Jang et al. \cite{jang_eum_2025} showed that while skin conductance level (SCL), skin conductance response (SCR), and blood volume pulse (BVP) were the most reliable indicators of pain, respiration rate also exhibited a significant decrease between the no-pain and pain states, suggesting it could serve as a valuable characteristic.
Finally, the authors in \cite{wu_chung_2012} noted that respiration is strongly modulated by emotion; however, raw traces are riddled with motion artifacts and mixed-emotion periods, which limit the reliability of automatic affect recognition. They introduced a parameter-free Respiration Quasi-Homogeneity Segmentation (RHS) algorithm to discard noisy segments, attaining high performance for affective (though not pain-related) states.

\section{Methodology}
This section describes the signal pre-processing steps, the architecture of the proposed model, the windowing strategies applied to the signal, and the fusion of features extracted from different windows using a gating mechanism for the final assessment. It also presents details on the augmentation and regularization techniques employed.

\subsection{Model Architecture}
\label{architecture}
The proposed model is a single cross-attention transformer called \textit{Resp-Encoder}, developed to extract a fixed-size representation from a respiration waveform. Designed for computational efficiency, the model employs a single cross-attention mechanism for global temporal aggregation, followed by a feed-forward refinement and a projection to a compact embedding space.
The input to the model is a respiration signal of duration $\theta$ seconds, represented as a sequence $\mathbf{r} \in \mathbb{R}^{\theta \times f \times 1}$, where $f$ denotes the sampling frequency in Hz.
To encode temporal structure, each time index is enriched with Fourier positional features. Specifically, sinusoidal basis functions with $K = 6$ frequency bands are applied up to a maximum frequency of 10\,Hz, resulting in a position-enhanced sequence $\tilde{\mathbf{r}} \in \mathbb{R}^{\theta \times d_{\text{in}}}$, where $d_{\text{in}} = 1 + 2K + 1$.
A set of $N = 256$ learnable latent vectors $\mathbf{L} \in \mathbb{R}^{N \times d}$, with $d = 512$, acts as a query bank. These latent vectors attend to the input sequence through a single-head cross-attention operation. The mechanism is intentionally asymmetric: the queries $\mathbf{Q} \in \mathbb{R}^{N \times d}$ are derived from the latent array, while the keys and values $\mathbf{K}, \mathbf{V} \in \mathbb{R}^{\theta \times d_{\text{in}}}$ are computed from the positionally encoded input. Since typically $N \ll \theta$, this configuration allows the model to efficiently summarize global input context without the computational burden of pairwise attention among all tokens.
The updated latent matrix $\mathbf{L}' \in \mathbb{R}^{N \times d}$ is processed by a gated feed-forward network (FFN) with residual connections and layer normalization. As the model consists of a single attention layer (depth $= 1$), this step provides the only non-linear transformation following attention. The resulting representation $\mathbf{L}'' \in \mathbb{R}^{N \times d}$ is mean-pooled across the latent dimension, yielding a single vector $\mathbf{e}_r \in \mathbb{R}^{d}$.
Finally, a linear projection maps this vector to a fixed-size output $\mathbf{z}_r \in \mathbb{R}^{512}$, which serves as the respiration embedding. The full transformation can be expressed as a mapping
\begin{equation}
\mathbf{r} \in \mathbb{R}^{\theta} \longrightarrow \mathbf{z}_r \in \mathbb{R}^{512}.
\end{equation}
Figure 1 presents an overview of the encoder architecture.
This architecture offers a balance between representational capacity and computational cost by combining global attention-based context modeling with a lightweight structure and minimal parameter count. A later section of the paper presents comparisons across different versions and configurations of the model, evaluating both predictive performance and efficiency.

\begin{figure}
\begin{center}
\includegraphics[scale=0.75]{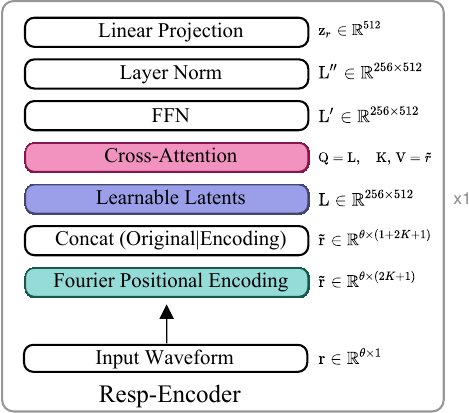}
\end{center}
\caption{Schematic overview of the proposed encoder.}
\label{signal}
\end{figure}

\subsection{Signal Pre-processing, Windowing \& Fusion}
To ensure a clean input, the respiratory signals were filtered using a $0.05-0.5$ $Hz$ band-pass filter, a range known to cover typical adult breathing frequencies  \cite{houtveen_groot_2006}, while effectively eliminating slow baseline drift and high-frequency cardiac or motion artifacts.
Figure \ref{signal} shows an example of a raw respiration signal alongside its filtered version.
Each respiration sequence is segmented into non-overlapping windows of duration $\theta = 5$ seconds. These fixed-length windows are treated as independent inputs to the model. If the final portion of the signal does not fully occupy a window, it is zero-padded to maintain consistent dimensions across samples. These fixed-size windows serve as independent inputs to the model in subsequent stages.
After windowing, each 5-second segment is independently processed by the \textit{Resp-Encoder} to obtain window-level embeddings:
\begin{equation}
\mathbf{z}_i \in \mathbb{R}^{512}, \qquad i = 1,\dots,S,
\end{equation}
where $S$ is the number of windows extracted from a signal.
To integrate information across windows, two representations are derived from these embeddings.  
An additive representation is computed by summing all window embeddings:
\begin{equation}
\mathbf{z}_{\text{add}} = \sum_{i=1}^{S} \mathbf{z}_i \in \mathbb{R}^{512},
\end{equation}
while a second representation is formed by concatenating them along the channel dimension:
\begin{equation}
\mathbf{z}_{\text{concat}} = \bigl[\mathbf{z}_1\,\Vert\,\dots\,\Vert\,\mathbf{z}_S\bigr] \in \mathbb{R}^{S \cdot 512}.
\end{equation}
In parallel, the complete respiration signal is passed through the same encoder to produce a full-sequence embedding:
\begin{equation}
\mathbf{z}_{\text{full}} \in \mathbb{R}^{512}.
\end{equation}

\begin{figure}
\begin{center}
\includegraphics[scale=0.275]{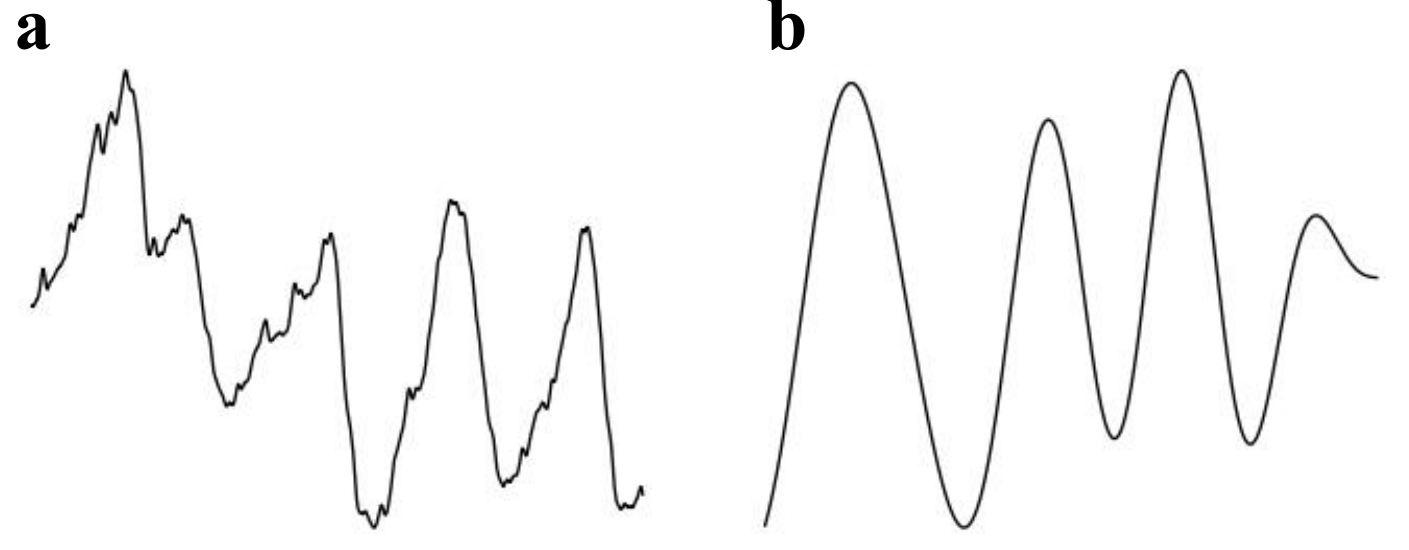}
\end{center}
\caption{Visualization of a respiration signal: (a) raw signal, (b) filtered signal.}
\label{signal}
\end{figure}

\subsection{Gate Mechanism}
\label{gate_mechanism}
\begin{figure*}
\begin{center}
\includegraphics[scale=1.26]{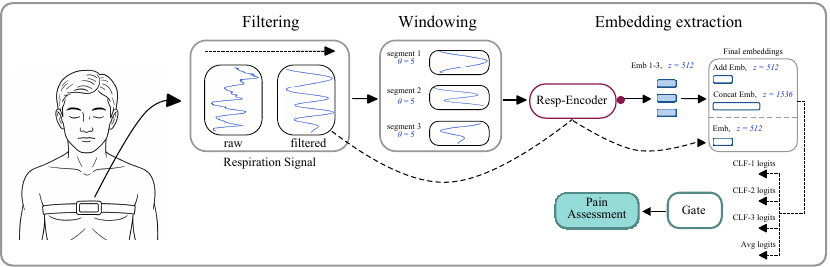}
\end{center}
\caption{Schematic overview of the proposed pipeline for pain assessment using respiration signals.}
\label{pipeline}
\end{figure*}

The pipeline produces four predictions from the three distinct input representations. Specifically, the additive fusion $\mathbf{z}_{\text{add}}$, the concatenated fusion $\mathbf{z}_{\text{concat}}$, and the full-signal embedding $\mathbf{z}_{\text{full}}$ are each passed by a dedicated one-layer classifier, resulting to their respective logits:
\begin{equation}
\mathbf{l}_{\text{add}},\qquad \mathbf{l}_{\text{concat}},\qquad \mathbf{l}_{\text{full}} \in \mathbb{R}^{C},
\end{equation}
where $C$ is the number of output classes.  
A fourth prediction is obtained by averaging the three logits:
\begin{equation}
\mathbf{l}_{\text{avg}} = \tfrac13\bigl(\mathbf{l}_{\text{add}} + \mathbf{l}_{\text{concat}} + \mathbf{l}_{\text{full}}\bigr).
\end{equation}
To select among these four logit sets on a \emph{per-sample} basis, a lightweight gating module is introduced.  
It uses a learnable parameter vector $\mathbf{g} \in \mathbb{R}^{4}$ to produce a one-hot weight vector for each sample, obtained via a hard Gumbel-Softmax:
\begin{equation}
\mathbf{w} = \text{Gumbel}_{\text{hard}}(\mathbf{g},\ \tau) \in \{0,1\}^{4}, \qquad \sum_{i=1}^{4} w_i = 1.
\end{equation}
The final logits for each sample are computed as:
\begin{equation}
\mathbf{l}_{\text{final}} =
      w_1\,\mathbf{l}_{\text{add}}
    + w_2\,\mathbf{l}_{\text{concat}}
    + w_3\,\mathbf{l}_{\text{full}}
    + w_4\,\mathbf{l}_{\text{avg}}.
\end{equation}
This gating mechanism assigns a one-hot weight vector to each candidate prediction, selecting one of them per sample. 
It enables the pipeline to adaptively select the most valuable output among different representations: local through the addition and concatenation of them, global through the full sequence, and the combination of all of them.
Figure \ref{pipeline} presents an overview of the proposed pipeline.

\subsection{Augmentation Methods \& Regularization}
Three data augmentation techniques are applied during training. Each method operates directly on the full-length respiration signal before any subsequent processing, including the windowing step described earlier.
First, signal \textit{Polarity inversion} multiplies the waveform by $-1$, flipping it across the horizontal axis.  
Second,
\textit{Gaussian noise} is added, where the signal-to-noise ratio (SNR) is randomly sampled from a range defined by:
\begin{equation}
\text{SNR} \in \bigl[0.001 \cdot k,\ 0.005 \cdot k\bigr], \quad k \sim \mathcal{U}(1,\ 1000).
\end{equation}
Finally, a \textit{Contiguous block masking} technique is applied, covering $10$–$30\,\%$ of the signal and masking it (set to zero). The block location is randomly chosen to be at the beginning, center, or end of the sequence with equal probability.
In addition, \textit{Dropout}, \textit{Label Smoothing}, as well as learning rate \textit{Warmup} and \textit{Cooldown} schedules are employed as regularization techniques.
Unless stated otherwise, their values are set to $10\%$, $10\%$, $50$, and $10$, respectively
Throughout all experiments, the batch size is fixed to $32$ and the learning rate is set to $1\mathrm{e}{-4}$.

\section{Experimental Evaluation \& Results}
This study leverages the dataset released by the challenge organizers, which consists of respiratory recordings from $65$ participants. Data collection took place at the Human-Machine Interface Laboratory, University of Canberra, Australia, and is divided into $41$ training, $12$ validation, and $12$ testing subjects. 
Pain stimulation was induced using transcutaneous electrical nerve stimulation (TENS) electrodes positioned on the inner forearm and the back of the right hand. 
Two pain levels were measured: pain threshold---the minimum stimulus intensity perceived as painful (low pain), and pain tolerance---the maximum intensity tolerated before becoming unbearable (high pain). 
Respiratory activity was recorded via a sensor placed on the participant’s chest.
The signals have a frequency of $100$ Hz and a duration of approximately $10$ seconds.
We refer to \cite{ai4pain_2025,rojas_hirachan_2023} for a detailed description of the recording protocol and to \cite{ai4pain_2024} for information regarding the previous edition of the challenge.
The experiments presented in this study are conducted on the validation subset of the dataset, evaluated under a multi-class classification framework with three levels: No Pain, Low Pain, and High Pain.
The validation results are reported in terms of macro-averaged accuracy, precision, and F1 score. The final results of the testing set are also reported. We note that all experiments followed a deterministic setup, eliminating the effect of random initializations; thus, any performance differences arose strictly from the chosen optimization settings, modalities, or other intentional changes rather than chance. Refer to Listing \ref{lst:determinism} for the implementation details.\\

\begin{center}
\begin{minipage}{0.95\linewidth} 
\begin{lstlisting}[caption={Deterministic setup for reproducibility.}, label={lst:determinism}]
from pytorch_lightning.utilities.seed import seed_everything
seed_everything(seed=3407)
torch.backends.cudnn.deterministic = True
torch.backends.cudnn.benchmark = False
\end{lstlisting}
\end{minipage}
\end{center}

\subsection{Architectural Components}
In the context of the model design, the number and size of its main components are evaluated, as efficiency is one of the primary objectives of the study.
As described in \ref{architecture}, the proposed configuration employs a single-layer cross-attention transformer. However, several alternative architectures have also been explored and evaluated to assess their performance and computational cost.
Table \ref{table:model_size} summarizes the corresponding results, while Table \ref{table:computational_cost} presents the computational cost for different module configurations in terms of millions of parameters and floating-point operations.
The experiments are based on six module configurations: (1) a model with one block containing one cross-attention module; (2) two consecutive blocks with one cross-attention module; (3) one block with one cross- and one self-attention module; (4) one block with one cross- and two self-attention modules; (5) two blocks with one cross- and one self-attention module; and (6) two blocks with one cross- and two self-attention modules. 
In all cases where both cross- and self-attention modules are present within a block, the self-attention module(s) are applied immediately after the corresponding cross-attention module in a consecutive manner.
These configurations exhibit an increasing trend in computational cost, ranging from the most efficient, with $3.62$ million parameters and $1.65$ GFLOPs, to the most complex, with $23.64$ million parameters and $11.88$ GFLOPs.
Three different epoch settings---$300$, $1200$, and $2100$---were also tested to evaluate the training duration required to achieve peak performance, as well as the model's behavior in terms of overfitting and generalization.

\begin{figure}
\begin{center}
\includegraphics[scale=0.235]{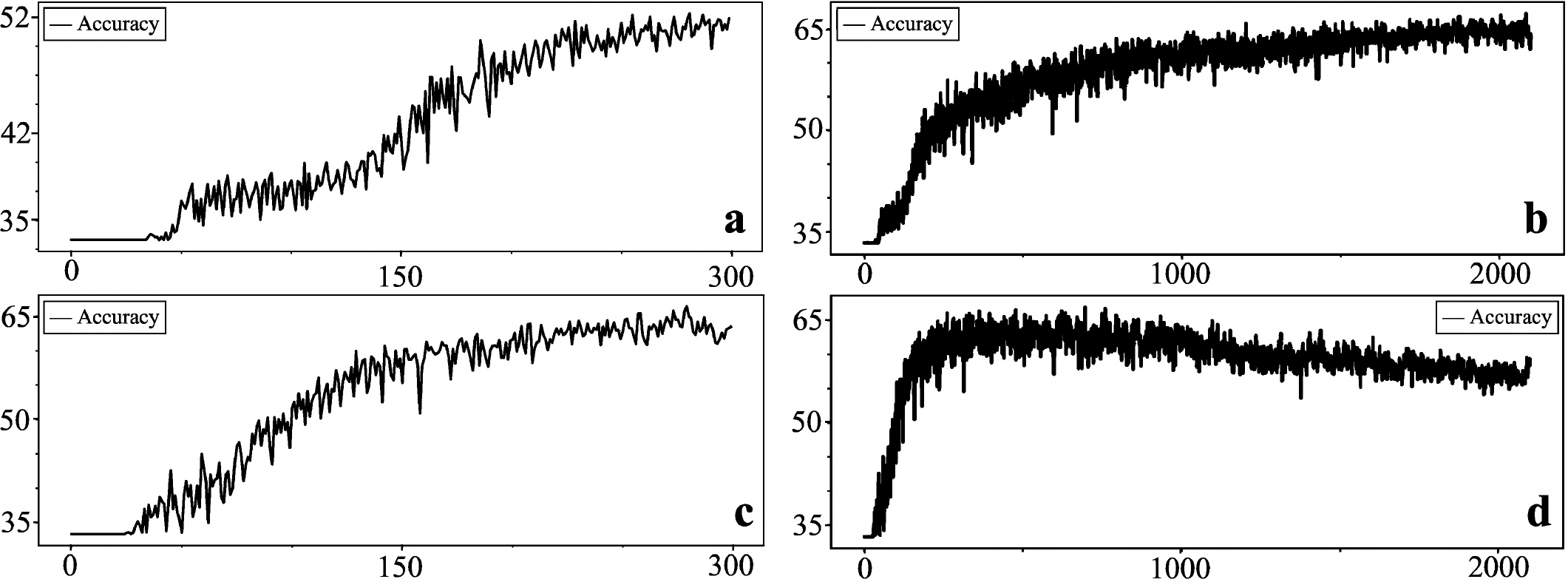}
\end{center}
\caption{Validation accuracies for \textit{1-1-0} at (a) 300 and (b) 2100 epochs, and \textit{2-1-2} at (c) 300 and (d) 2100 epochs; the heavier \textit{2-1-2} peaks near epoch 800 and then declines, whereas the lighter \textit{1-1-0} continues to improve without overfitting.}
\label{validation_size_modules}
\end{figure}

\begin{figure}
\begin{center}
\includegraphics[scale=0.0593]{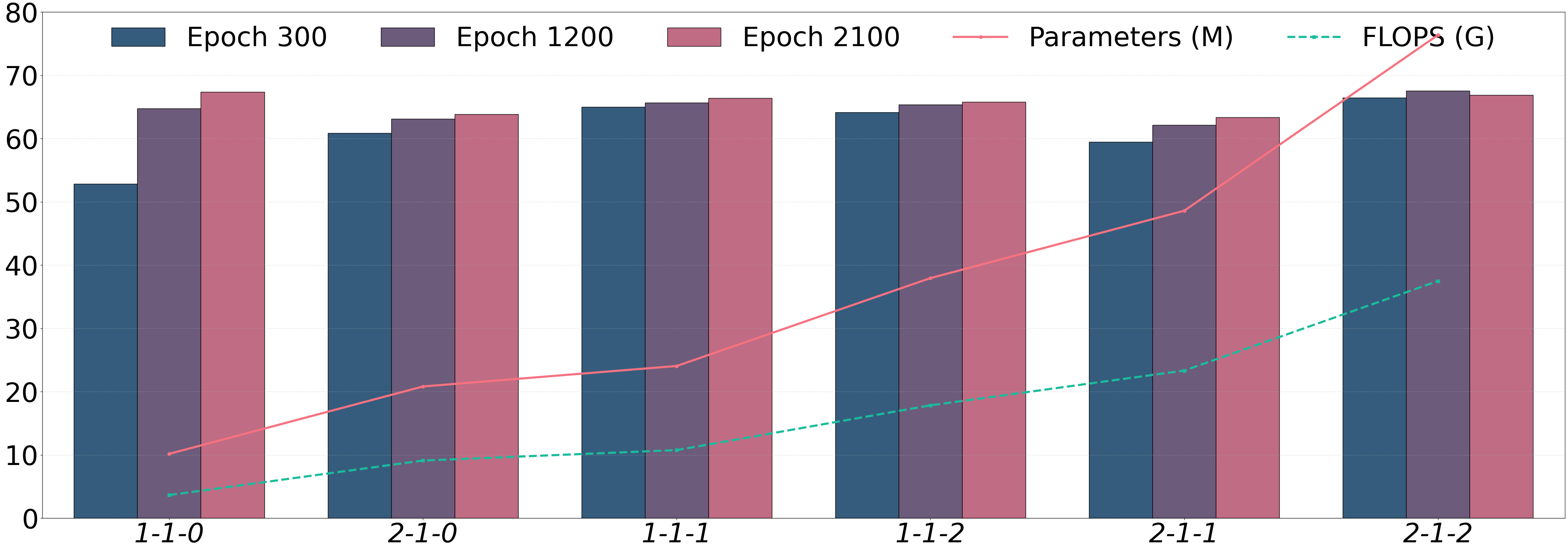}
\end{center}
\caption{Overview of accuracy, parameter count, and computational cost across different module configurations and training durations.}
\label{cost_comparison}
\end{figure}

We observe that the most efficient configuration, denoted as \textit{1-0-0}, achieved an accuracy of $52.84\%$ after $300$---epochs substantially lower than other configurations, particularly the largest model, \textit{2-1-2}, which reached $66.47\%$. Interestingly, the \textit{1-0-0} setup yielded one of the highest precision scores at $67.78\%$, only slightly behind the $68.19\%$ of 1-1-1 and $70.79\%$ of \textit{2-1-2}. Regarding the F1-score, the trend followed that of accuracy, indicating that model size is directly related to performance at this stage.
When the number of epochs was increased to $1200$, the \textit{1-1-0} configuration exhibited a significant improvement of over $10\%$, reaching an accuracy of $64.73\%$. In contrast, the remaining configurations saw minimal gains---mostly around $2\%$. For instance, \textit{2-1-0} improved from $60.86\%$ to $63.10\%$, \textit{1-1-2} from $64.13\%$ to $65.38\%$, and \textit{2-1-2} reached $67.57\%$, suggesting that the larger models may have already approached a performance plateau. Precision for \textit{1-1-0} increased to $71.71\%$, while other configurations showed a decline---for example, \textit{2-1-1} and \textit{2-1-2} dropped by $1.36$ and $1.28$ points, respectively. F1-scores slightly increased for most configurations, except \textit{1-1-0}, which showed a substantial jump of $12.13\%$.
Extending training to $2100$ epochs further improved performance for \textit{1-1-0}, which reached $67.33\%$ accuracy---second only to \textit{2-1-2}'s $67.57\%$. It also achieved the highest precision ($73.74\%$) and F1-score ($69.95\%$) across all configurations and metrics.
These results indicate that the largest models do not necessarily yield the best performance. On the contrary, smaller configurations demonstrated strong and often superior outcomes. As previously noted, larger models tend to reach their performance ceiling earlier, whereas smaller ones continue to improve—a pattern observed across all proposed configurations in this study. Figure \ref{validation_size_modules} illustrates the validation performance of \textit{1-1-0} and \textit{2-1-2} across $300$ and $2100$ epochs. It is evident that the larger model peaks around $800$ epochs and subsequently suffers from performance degradation due to overfitting. In contrast, \textit{1-1-0}---while underperforming at $300$ epochs—shows no overfitting even at $2100$ epochs and continues to exhibit potential for further improvement.
Finally, Figure \ref{cost_comparison} illustrates the performance trends across module configurations, training epochs, and computational cost---namely, the number of parameters and FLOPS. Notably, at $2100$ epochs, the \textit{1-1-0} configuration achieves nearly identical accuracy to \textit{2-1-2} while requiring over seven times fewer FLOPS and twenty times fewer parameters.
The next series of experiments are based on the \textit{1-1-0} configuration, chosen for its strong performance and efficiency.

\begin{table}
\caption{Comparison of performances for different module configurations.}
\label{table:model_size}
\begin{center}
\begin{threeparttable}
\begin{tabular}{ P{1.0cm} P{0.65cm} P{0.65cm}  P{0.65cm}  P{1.2cm} P{1.2cm} P{0.6cm}  }

\toprule
\multirow{2}[2]{*}{\shortstack{Epochs}}
&\multicolumn{3}{c}{Architecture} 
&\multicolumn{3}{c}{Task--MC}\\ 
\cmidrule(lr){2-4}\cmidrule(lr){5-7}
&Depth &Cross &Self &Accuracy &Precision &F1\\
\midrule
\midrule
300   &1 &1 &-- &52.84             &67.78          &55.54   \\
1200  &1 &1 &-- &64.73             &71.71          &67.67   \\
2100  &1 &1 &-- &\underline{67.33} &\textbf{73.74} &\textbf{69.95}   \\\hdashline

300   &2 &1 &-- &60.86 &65.95 &63.01   \\
1200  &2 &1 &-- &63.10 &64.48 &63.59   \\
2100  &2 &1 &-- &63.86 &67.68 &65.21   \\\hdashline

300   &1 &1 &1 &64.99 &68.19 &66.32   \\
1200  &1 &1 &1 &65.63 &69.33 &67.07   \\
2100  &1 &1 &1 &66.39 &68.03 &67.17   \\\hdashline

300   &1 &1 &2 &64.13 &66.62 &65.23   \\
1200  &1 &1 &2 &65.38 &66.48 &65.07   \\
2100  &1 &1 &2 &65.75 &66.94 &66.62   \\\hdashline

300   &2 &1 &1 &59.44 &67.00 &62.42   \\
1200  &2 &1 &1 &62.11 &65.36 &63.20   \\
2100  &2 &1 &1 &63.37 &67.94 &65.07   \\\hdashline

300   &2 &1 &2 &66.47          &\underline{70.79} &68.39   \\
1200  &2 &1 &2 &\textbf{67.57} &69.51             &\underline{68.49}   \\
2100  &2 &1 &2 &66.87          &69.98             &67.63   \\
\bottomrule 
\end{tabular}
\begin{tablenotes}[para,flushleft] 
\scriptsize                   
\item \textbf{Depth}: number of stacked \texttt{[cross/self]} blocks; e.g.\ Depth = 2 means two consecutive \texttt{[cross/self]} layers
\textbf{Cross / Self}: number of cross- and self-attention modules per block
\textbf{MC}: multiclass classification (No, Low, High Pain)\
\textbf{Bold}: best performance per metric. \underline{Underline}: second-best performance.
\end{tablenotes}
\end{threeparttable}
\end{center}
\end{table}

\begin{table}
\caption{Number of parameters and FLOPS for different modules configurations.}
\label{table:computational_cost}
\begin{center}
\begin{threeparttable}
\begin{tabular}{ P{0.7cm} P{0.7cm} P{0.7cm} P{2.0cm} P{2.0cm} }
\toprule
\multicolumn{3}{c}{Architecture} 
&\multicolumn{2}{c}{Computational Cost}\\
\cmidrule(lr){1-3}\cmidrule(lr){4-5}
Depth & Cross & Self &Parameters (M) &FLOPS (G)\\              
\midrule
\midrule
1 & 1 & -- &3.62  &1.65  \\
2 & 1 & -- &6.84  &3.30  \\
1 & 1 & 1  &7.82  &3.80  \\
1 & 1 & 2  &12.02 &5.94  \\
2 & 1 & 1  &15.24 &7.60  \\
2 & 1 & 2  &23.64 &11.88 \\
\bottomrule
\end{tabular}
\begin{tablenotes}
\scriptsize
\item \textbf{M}: millions \textbf{G}: giga
\end{tablenotes}
\end{threeparttable}
\end{center}
\end{table}


\subsection{Signal Padding}
\label{signal_padding}
\begin{figure}
\begin{center}
\includegraphics[scale=0.052]{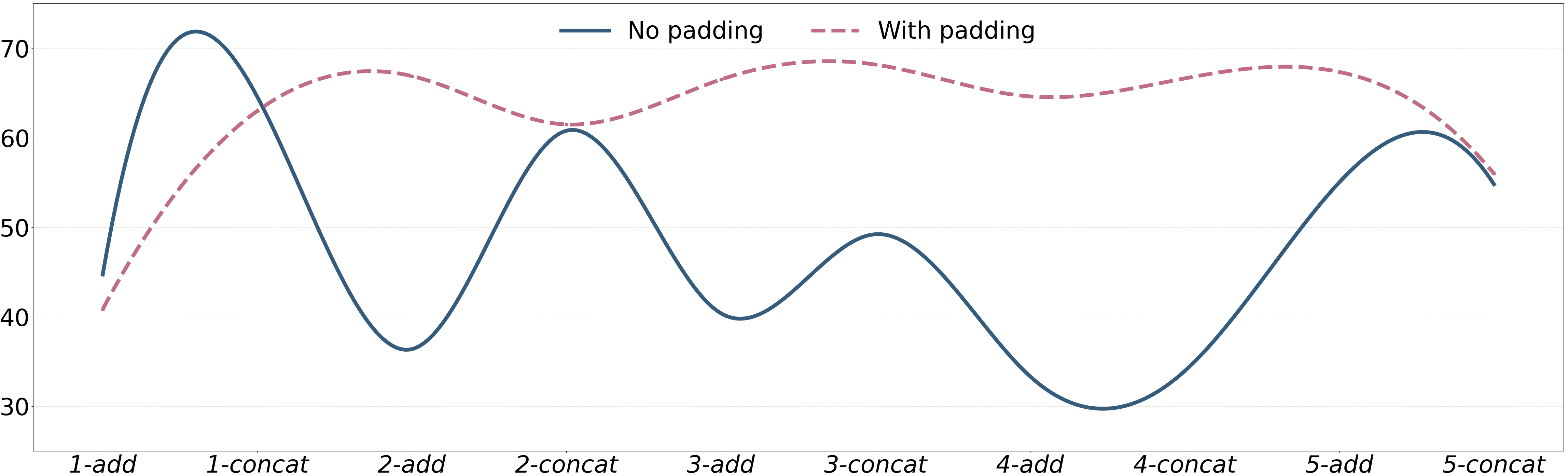}
\end{center}
\caption{Accuracy comparison across different window durations with and without padding.}
\label{padding}
\end{figure}

This section focused on evaluating the effect of signal padding as the main experimental factor, examining how it influences both window duration and fusion strategy in respiratory signal analysis. Experiments were conducted over $300$ epochs using five different window durations, denoted as $T$: $T = 1$, $2$, $3$, $4$, and $5$ seconds.
Two fusion approaches---addition and concatenation---were applied to features extracted from each window. 
We note that a single \textit{Resp-Encoder} is used for all signal segments.
Table \ref{table:padding} presents the corresponding results.
All signals have a sampling frequency of $100 Hz$ and a duration of around $10$ seconds, resulting in vectors of approximately $1000$ data points. To standardize the input length, we apply zero-padding to extend each vector to a fixed length of $1150$ data points.
Without applying padding and using additive fusion, performance ranges from $44.72\%$ for $1$-second windows to $55.03\%$ for $5-$second windows. A notable training collapse is observed at $T=4$, with performance dropping to $33.33\%$---equivalent to random choice. When padding is applied, performance improves substantially across almost all window durations, with the highest accuracy reaching $67.36\%$ at $T=5$. Additionally, the collapse observed at $T=4$ no longer occurs.
For the concatenation fusion method, we observe a similar pattern. Without padding, the average accuracy is $52.67\%$, while padding increases it to $63.06\%$, a gain of more than $10$ percentage points. The highest performance of $68.18\%$ is achieved at $T=3$. Again, a learning collapse occurs at $T=4$ without padding, but this issue is resolved when padding is used, resulting in an accuracy of $66.64\%$.
Figure \ref{padding} illustrates the performance trends with and without the padding mechanism, clearly showing the consistent improvements. Given these results, the padding strategy will be adopted as the default configuration in subsequent experiments.

\begin{table}
\caption{Performance comparison across different window durations (\textit{T}), fusion strategies (addition \& concatenation), and the effect of applying zero-padding.}
\label{table:padding}
\begin{center}
\begin{threeparttable}
\begin{tabular}{ P{0.9cm} P{0.3cm} P{0.9cm} P{1.0cm} P{1.1cm} P{1.1cm} P{0.7cm}  }

\toprule
\multirow{2}[2]{*}{\shortstack{Epochs}}
&\multicolumn{3}{c}{Windowing} 
&\multicolumn{3}{c}{Task--MC}\\ 
\cmidrule(lr){2-4}\cmidrule(lr){5-7}
&$T$ &Fusion &Padding &Accuracy &Precision &F1\\
\midrule
\midrule
300 &1 &add    &--         &44.72             &58.54             &45.34  \\
300 &2 &add    &--         &36.38             &60.74             &35.70  \\
300 &3 &add    &--         &40.38             &60.21             &41.01  \\
300 &4 &add    &--         &33.33             &26.78             &29.70  \\
300 &5 &add    &--         &55.03             &60.28             &57.15  \\\hdashline
300 &1 &add    &\checkmark &40.84             &41.93             &40.39  \\
300 &2 &add    &\checkmark &\underline{66.90} &\textbf{71.31}    &64.45  \\
300 &3 &add    &\checkmark &66.52             &68.61             &\underline{67.36}  \\
300 &4 &add    &\checkmark &64.62             &66.67             &58.76  \\
300 &5 &add    &\checkmark &\textbf{67.36}    &\underline{69.27} &\textbf{67.91}  \\\midrule
300 &1 &concat &--         &64.59             &\textbf{71.36}    &63.62  \\
300 &2 &concat &--         &60.80             &66.71             &63.15  \\
300 &3 &concat &--         &49.23             &63.12             &52.80  \\
300 &4 &concat &--         &33.94             &43.48             &31.02  \\
300 &5 &concat &--         &54.81             &64.07             &57.89  \\\hdashline
300 &1 &concat &\checkmark &63.00             &65.53             &63.99  \\
300 &2 &concat &\checkmark &61.49             &67.91             &64.09  \\
300 &3 &concat &\checkmark &\textbf{68.18}    &69.84             &\textbf{68.67}  \\
300 &4 &concat &\checkmark &\underline{66.64} &\underline{70.07} &\underline{67.91}  \\
300 &5 &concat &\checkmark &56.00             &65.92             &59.47  \\

\bottomrule 
\end{tabular}
\begin{tablenotes}[para,flushleft] 
\scriptsize                   
\item \textbf{$T$}: duration of each window in seconds
\end{tablenotes}
\end{threeparttable}
\end{center}
\end{table}

\subsection{Windows Size}
\begin{figure}
\begin{center}
\includegraphics[scale=0.0539]{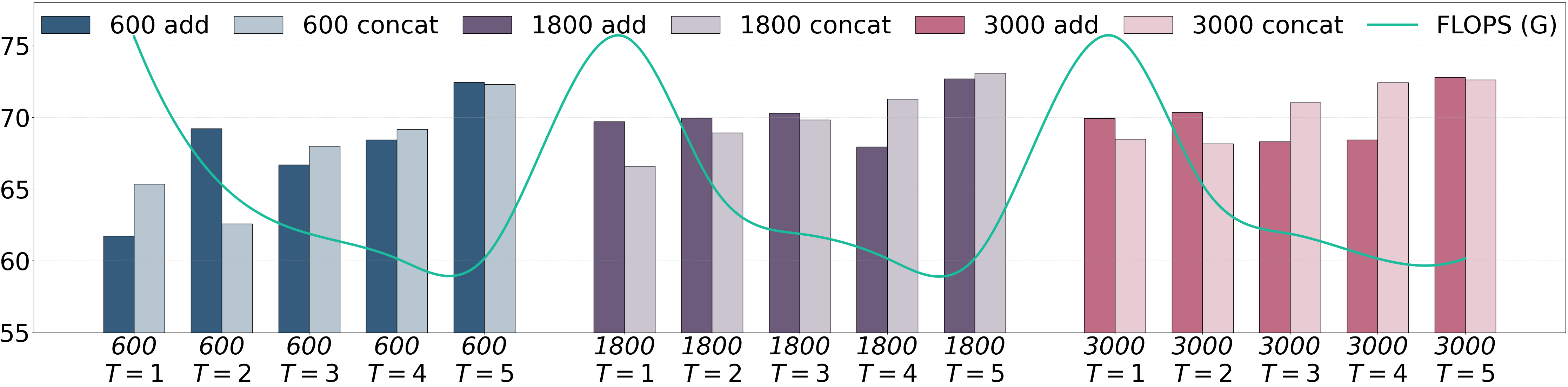}
\end{center}
\caption{Comparison of classification accuracy and computational cost across different training durations, window sizes (\textit{T}), and fusion methods. }
\label{windows_fusion}
\end{figure}

Building on the results in Section \ref{signal_padding}, we further optimize the fusion of different window types with respect to window size. The \textit{1-1-0} configuration, being lightweight, requires a more extended training period; however, previous experiments have shown that it exhibits stable learning and strong performance. In this section, we investigate how increasing the number of training epochs influences the learning dynamics, explore the peak performance achievable for each window size and fusion approach, and identify when overfitting begins to occur. Table \ref{table:windows_fusion} presents the corresponding results.
We observe that the smallest window size of $1$ is more influenced by training duration. At $600$ epochs with $T=1$, the accuracy reaches $61.72\%$ and $65.34\%$ for addition and concatenation fusion, respectively. As training progresses to $1800$ epochs, these values rise to $69.72\%$ and $66.61\%$ and further to $69.94\%$ and $68.49\%$ at $3000$ epochs. In contrast, the performance gains for longer windows are minor. For $T=5$, the accuracy with addition increases only slightly from $72.45\%$ at $600$ epochs to $72.69\%$ and $72.80\%$ at $1800$ and $3000 epochs$, respectively. A similar trend is observed for concatenation. These results suggest that performance plateaus as window size increases, with $T=5$ consistently achieving the best results among all configurations.
It is also important to note the difference in computational cost across window sizes. Smaller windows require more segments to cover the full signal, resulting in higher overall computational costs, while longer windows reduce the number of segments needed.
Figure \ref{windows_fusion} illustrates the relationship between the number of training epochs, window size, fusion method, and computational cost. As discussed, $T=5$ emerges as the most effective window size, yielding the best accuracy while maintaining the lowest computational cost---only $4.94$ GFLOPs. Regarding fusion methods, both addition and concatenation exhibit similar behavior, achieving peak accuracies of $72.80\%$ and $73.09\%$, respectively, with no significant differences.
Note that all the following experiments are based on $T=5$.

\begin{table}
\caption{Performance comparison across different window durations (\textit{T}), fusion strategies (addition \& concatenation), and computational cost.}
\label{table:windows_fusion}
\begin{center}
\begin{threeparttable}
\begin{tabular}{ P{0.8cm} P{0.3cm} P{0.9cm} P{1.35cm}  P{1.05cm} P{0.91cm} P{0.61cm}  }

\toprule
\multirow{2}[2]{*}{\shortstack{Epochs}}
&\multicolumn{3}{c}{Windowing} 
&\multicolumn{3}{c}{Task--MC}\\ 
\cmidrule(lr){2-4}\cmidrule(lr){5-7}
&$T$ &Fusion &FLOPS {\footnotesize(\text{G})} &Accuracy &Precision &F1\\
\midrule
\midrule
600 &1 &add    &19.74   &61.72              &62.64             &61.15  \\
600 &2 &add    &9.87    &\underline{69.21}  &70.48             &68.70  \\
600 &3 &add    &6.58    &66.70              &68.44             &64.13  \\
600 &4 &add    &4.93    &68.43              &\underline{70.58} &\underline{69.06}  \\
600 &5 &add    &4.94    &\textbf{72.45}     &\textbf{74.28}    &\textbf{72.54}  \\\hdashline
600 &1 &concat &19.74   &65.34              &68.20             &66.53  \\
600 &2 &concat &9.87    &62.57              &65.48             &61.98  \\
600 &3 &concat &6.58    &68.00              &65.56             &68.08  \\
600 &4 &concat &4.93    &\underline{69.16}  &\underline{72.99} &\underline{69.47}  \\
600 &5 &concat &4.94    &\textbf{72.31}     &\textbf{73.94}    &\textbf{73.05}  \\\midrule

1800 &1 &add    &19.74  &69.72             &68.86 	           &66.05  \\
1800 &2 &add    &9.87   &69.95             &70.65 	           &\underline{69.99}  \\
1800 &3 &add    &6.58   &\underline{70.29} &\underline{71.91}  &69.93  \\
1800 &4 &add    &4.93   &67.94             &69.15 	   	       &67.45  \\
1800 &5 &add    &4.94   &\textbf{72.69}    &\textbf{75.45} 	   &\textbf{73.78}  \\\hdashline
1800 &1 &concat &19.74  &66.61             &68.49 	           &67.36  \\
1800 &2 &concat &9.87   &68.92             &72.89 	           &69.84  \\
1800 &3 &concat &6.58   &69.82             &70.42              &70.11  \\
1800 &4 &concat &4.93   &\underline{71.27} &\underline{73.44}  &\underline{72.21}  \\
1800 &5 &concat &4.94   &\textbf{73.09}    &\textbf{73.59} 	   &\textbf{73.31}  \\\midrule

3000 &1 &add    &19.74  &69.94             &70.44              &70.21  \\
3000 &2 &add    &9.87   &\underline{70.34} &\underline{70.93}  &\underline{70.41}  \\
3000 &3 &add    &6.58   &68.32             &70.32              &68.56  \\
3000 &4 &add    &4.93   &68.43             &69.71              &68.47  \\
3000 &5 &add    &4.94   &\textbf{72.80}    &\textbf{74.43}     &\textbf{72.03}  \\\hdashline
3000 &1 &concat &19.74  &68.49             &69.79              &69.11  \\
3000 &2 &concat &9.87   &68.17             &70.98              &69.33  \\
3000 &3 &concat &6.58   &71.04             &71.91              &71.45  \\
3000 &4 &concat &4.93   &\underline{72.43} &\textbf{74.74}     &\textbf{73.39}  \\
3000 &5 &concat &4.94   &\textbf{72.63}    &\underline{73.69}  &\underline{73.04}  \\

\bottomrule 
\end{tabular}
\begin{tablenotes}[para,flushleft] 
\scriptsize                   
\item 
\end{tablenotes}
\end{threeparttable}
\end{center}
\end{table}

\subsection{Fusion Strategies}
As previously discussed, the addition and concatenation of window-level embeddings yield similar performance with no substantial differences. To further explore their potential, we evaluated combinations of these two fusion methods along with the full-sequence representation (which also demonstrated strong performance without windowing). Unless stated otherwise, all experiments were conducted using $3000$ training epochs. The results are summarized in Table \ref{table:final_windows_fusion}.
The additive and concatenated window embeddings were first combined to form a joint representation,
$\mathbf{z}_{\text{add+concat}} = [\mathbf{z}_{\text{add}} \,\Vert\, \mathbf{z}_{\text{concat}}]$,
which achieved an accuracy of $72.37\%$.
Next, the full-sequence representation was incorporated by concatenating it with the previously fused vector:
$\mathbf{z}_{\text{all}} = [\mathbf{z}_{\text{add}} \,\Vert\, \mathbf{z}_{\text{concat}} \,\Vert\, \mathbf{z}_{\text{full}}]$,
resulting in a lower accuracy of $63.84\%$, approximately $9\%$ lower than the best-performing configuration.
Late fusion strategies were also explored. In one case, predictions were obtained independently from the fused embedding and the full-sequence embedding and then averaged:
$\mathbf{l}_{\text{avg}} = \tfrac{1}{2}(\mathbf{l}_{\text{fused}} + \mathbf{l}_{\text{full}})$,
yielding $65.84\%$ accuracy. Replacing the fixed average with a learnable scalar weight $\alpha \in [0, 1]$,
$\mathbf{l}_{\text{weighted}} = \alpha \mathbf{l}_{\text{fused}} + (1 - \alpha) \mathbf{l}_{\text{full}}$,
resulted in slightly lower performance at $65.14\%$.
Finally, the proposed gating mechanism described in \ref{gate_mechanism} was applied, which adaptively selects one of the four candidate predictions per sample. This approach achieved an accuracy of $64.76\%$.
It is important to note that incorporating the full-sequence signal did not directly enhance performance but rather contributed to a more stable learning process. A similar effect was observed when unifying the two types of window embeddings, where the learning curves were smoother and lacked the abrupt performance spikes seen in other configurations.
This behavior is likely due to the increased complexity introduced by combining embeddings from different sources, such as the windowed segments and the full signal, which makes the optimization process more challenging. However, this complexity appears to act as a form of regularization, promoting gradual and stable convergence during training.
Based on these observations, the subsequent experiments utilize the combination of all embeddings, including both windowed and full-sequence representations, via the proposed gating mechanism.
We also refer readers to the right part of Figure \ref{pipeline} for a more intuitive, visual understanding of how embeddings are derived from the signal segments and how the proposed fusion strategy is applied.

\begin{figure}
\begin{center}
\includegraphics[scale=0.17]{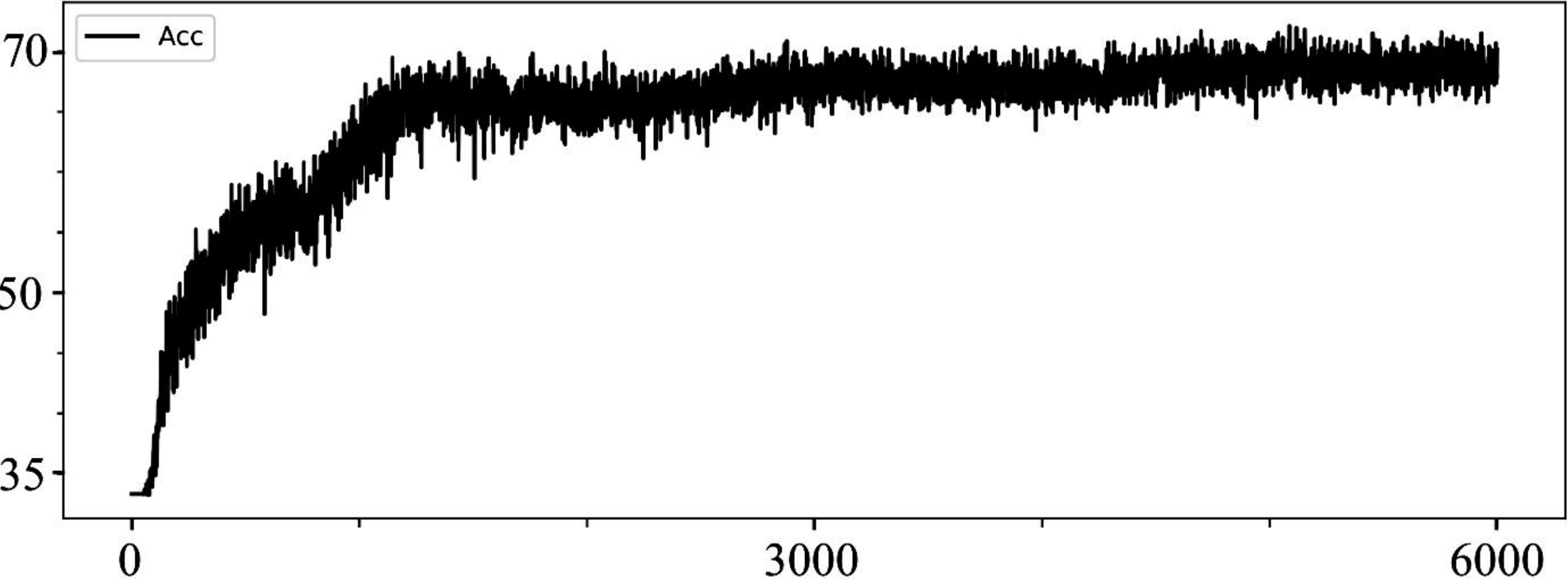}
\end{center}
\caption{Validation accuracy curve corresponding to the final training setup, combining all available inputs---windowed embeddings and the full signal---via the proposed gating mechanism, trained for 6000 epochs.}
\label{validation_final}
\end{figure}

Further experiments were conducted to adjust augmentation and regularization settings in order to optimize performance, as shown in Table \ref{table:final_optimization}.
Changing the augmentation probability from a fixed $50\%$ to a dynamic range of $0-100\%$ increased accuracy from $64.76\%$ to $65.05\%$. Lowering the probability to $20\%$ and increasing the dropout rate to $15\%$ resulted in a significant increase to $71.04\%$. Removing label smoothing and increasing dropout to $30\%$ led to an accuracy of $71.67\%$. 
Setting label smoothing to $10\%$ while raising dropout to $40\%$ reduced accuracy to $70.87\%$ but yielded smoother learning curves. 
Due to the observed training stability, the $10\%$ label smoothing and $40\%$ dropout configuration was extended to $6000$ epochs, achieving an accuracy of $72.12\%$ without indications of overfitting.
Figure \ref{validation_final} shows the validation curve for the corresponding training setup.

\begin{table}
\caption{Evaluation of fusion strategies combining window-level and full-sequence representations.
All experiments were conducted with 3000 training epochs. Metrics are reported as Accuracy\,\textbar\,Precision\,\textbar\,F1 (\%).}
\label{table:final_windows_fusion}
\begin{center}
\begin{threeparttable}
\begin{tabular}{P{1.70cm} P{1.1cm}  P{0.60cm}  P{1.60cm} P{1.7cm} }

\toprule
\multicolumn{2}{c}{Windowing} 
&\multicolumn{2}{c}{Extra} 
&\multicolumn{1}{c}{Task--MC}\\ 
\cmidrule(lr){1-2}\cmidrule(lr){3-4}\cmidrule(lr){5-5}
Input\textsuperscript{1} &Fusion\textsuperscript{1} &Input\textsuperscript{2} &Fusion\textsuperscript{2} &Metrics\\
\midrule
\midrule
w-add, w-cat  &concat &--   &-- &72.37\scriptsize{\textbar 72.20\textbar 72.21}  \\
w-add, w-cat  &concat &full &-- &63.84\scriptsize{\textbar 63.84\textbar 65.51}  \\\hdashline

w-add, w-cat  &concat &full & LF-avg   &65.84\scriptsize{\textbar 72.53 \textbar 68.33}  \\
w-add, w-cat  &concat &full & LF-coef  &65.14\scriptsize{\textbar 71.44 \textbar 67.44}  \\

w-add, w-cat  &concat &full &LF-avg-gate   &64.76\scriptsize{\textbar 69.50 \textbar 66.81}  \\

\bottomrule 
\end{tabular}
\begin{tablenotes}[para,flushleft] 
\scriptsize                   
\item \textbf{Input\textsuperscript{1}} and \textbf{Fusion\textsuperscript{1}} refer to the initial fusion of window-level embeddings: \textbf{\texttt{w-add}} denotes element-wise addition and \textbf{\texttt{w-cat}} denotes concatenation. \textbf{Input\textsuperscript{2}} refers to any additional input used, such as the full (unwindowed) signal. \textbf{Fusion\textsuperscript{2}} indicates the second-level combination method: \textbf{\texttt{LF-avg}} averages the logits from the fused window representation and the full signal; \textbf{\texttt{LF-coef}} uses a learnable scalar to weight their contribution; \textbf{\texttt{LF-avg-gate}} applies the proposed Gumbel-Softmax gating to select among all candidate outputs. 
\textbf{MC}: multiclass classification (No, Low, High Pain)
\end{tablenotes}
\end{threeparttable}
\end{center}
\end{table}

\begin{table}
\caption{Performance impact of different augmentation and regularization settings. Metrics are reported as Accuracy\,\textbar\,Precision\,\textbar\,F1 (\%).}

\label{table:final_optimization}
\begin{center}
\begin{threeparttable}
\begin{tabular}{P{1.0cm} P{0.90cm}  P{0.9cm} P{1.00cm}  P{0.60cm} P{1.7cm} }

\toprule
\multicolumn{3}{c}{Augmentations} 
&\multicolumn{2}{c}{Regularization} 
&\multicolumn{1}{c}{Task--MC}\\ 
\cmidrule(lr){1-3}\cmidrule(lr){4-5}\cmidrule(lr){6-6}
Polarity &Noise &Mask &LS &DO  &Metrics\\
\midrule
\midrule
50\textbar 50  &50\textbar 50 &50\textbar 50 &10 &10 &64.76\scriptsize{\textbar 69.50 \textbar 66.81}\\
0\textbar 100 &0\textbar 100  &0\textbar 100 &10 &10 &65.05\scriptsize{\textbar 71.52 \textbar 67.75}\\

20\textbar 20  &20\textbar 20 &20\textbar 20 &10 &15 &71.04\scriptsize{\textbar 72.84 \textbar 71.83}\\
20\textbar 20  &20\textbar 20 &20\textbar 20 &0  &30 &71.67\scriptsize{\textbar 73.14 \textbar 72.38}\\

20\textbar 20  &20\textbar 20 &20\textbar 20 &10 &40 &70.87\scriptsize{\textbar 71.47 \textbar 71.00}\\\midrule
\rowcolor{mygray}20\textbar 20  &20\textbar 20 &20\textbar 20 &10 &40 &72.12\scriptsize{\textbar 71.86 \textbar 71.87}\\

\bottomrule 
\end{tabular}
\begin{tablenotes}[para,flushleft] 
\scriptsize                   
\item Values in the format \textbf{$x\textbar y$} represent sample-wise augmentation probabilities randomly drawn from the range $[x\%, y\%]$ during training. \textbf{LS}: label smoothing (in \%) \textbf{DO}:dropout rate (in \%). 
\raisebox{0.15ex}{\colorbox{mygray}{\phantom{\rule{0.6em}{1ex}}}}: indicates the model trained for $6000$ epochs

\end{tablenotes}
\end{threeparttable}
\end{center}
\end{table}


\section{Comparison with Existing Methods}
In this section, the proposed approach is compared with previous studies using the testing set of the \textit{AI4PAIN} dataset. Some of these studies were conducted as part of the \textit{First Multimodal Sensing Grand Challenge}. In contrast, others, including the present work, used data from the \textit{Second Multimodal Sensing Grand Challenge}---the main distinction between the two lies in the availability of different modalities.
Studies employing facial video or fNIRS have reported strong results, with accuracies of $49.00\%$ by \cite{prajod_schiller_2024} and $55.00\%$ by \cite{nguyen_yang_2024}, respectively. The combination of these two modalities also yielded high results, although not significantly higher than when each modality was used in isolation.
For example, \cite{vianto_2025} reported $51.33\%$, and \cite{gkikas_rojas_painformer_2025} achieved $55.69\%$ using fused video and fNIRS data.
Concerning the physiological modalities available in the \textit{Second Grand Challenge}, even higher accuracies have been reported. In \cite{gkikas_kyprakis_multimodal_2025}, the authors reached $54.89\%$ using a combination of EDA, BVP, respiration, and 
blood oxygen saturation (SpO$_2$), while in \cite{gkikas_kyprakis_eda_2025} achieved $55.17\%$ using only EDA.
The proposed method, based solely on respiration signals, achieved an accuracy of $42.24\%$. While this result is lower than those reported in studies using more or different modalities either in isolation or in combination, it is consistent with known limitations of respiration as a single-modality source in pain recognition.

\begin{table}
\caption{Comparison of studies on the testing set of the \textit{AI4Pain} dataset.}
\label{table:ai4pain_test}
\begin{center}
\begin{threeparttable}
\begin{tabular}{P{0.7cm} P{2.0cm} P{3.0cm} P{1.0cm}}
\toprule
Study & Modality & ML & Acc (\%) \\
\midrule
\midrule
\cite{khan_aziz_2025}$^\dagger$                &fNIRS        &ENS             &53.66\\ \hdashline
\cite{nguyen_yang_2024}$^\dagger$              &fNIRS        &Transformer     &55.00\\ \hdashline
\cite{prajod_schiller_2024}$^\dagger$          &Video        &2D CNN          &49.00\\ \hdashline
\cite{gkikas_tsiknakis_painvit_2024}$^\dagger$ &Video, fNIRS &Transformer     &46.67\\ \hdashline
\cite{vianto_2025}$^\dagger$                   &Video, fNIRS &CNN-Transformer &51.33 \\\hdashline
\cite{gkikas_rojas_painformer_2025}$^\dagger$  &Video, fNIRS &Transformer     &55.69 \\\midrule

\cite{gkikas_kyprakis_multimodal_2025}$^\ddagger$    &EDA, BVP, Resp, SpO$_2$ &MoE     &54.89\\ \hdashline
\cite{gkikas_kyprakis_eda_2025}$^\ddagger$    &EDA                     &Transformer &55.17\\\hdashline
Our$^\ddagger$        &Respiration             &Transformer &42.24 \\

\bottomrule 
\end{tabular}
\begin{tablenotes}
\scriptsize
\item \textbf{ENS}: Ensemble Classifier  \textbf{SpO$_2$}: Peripheral Oxygen Saturation \textbf{MoE}: Mixture of Experts 
$\pmb{\dagger}$: AI4PAIN-First Multimodal Sensing Grand Challenge $\pmb{\ddagger}$: AI4PAIN-Second Multimodal Sensing Grand Challenge
\end{tablenotes}
\end{threeparttable}
\end{center}
\end{table}

\section{Discussion \& Conclusion}
This study presents our contribution to the \textit{Second Multimodal Sensing Grand Challenge for Next-Generation Pain Assessment (AI4PAIN)}, where respiration signals were the chosen modality.
With respect to the model, we developed an efficient transformer-based architecture that employs a single cross-attention mechanism. Experimental results show that this compact model not only outperforms heavier counterparts but also achieves markedly lower computational cost and higher efficiency---factors that, in the current era of AI and deep learning, researchers must carefully consider and value.
Regarding the proposed pipeline, a multi-window-based approach was introduced to extract information from local regions of the signal and fuse the corresponding embeddings in various ways. Additionally, incorporating the original signal sequence, beyond retaining global information, contributed to more stable learning during training.
The results indicated solid performance, particularly after optimization. However, the final test set results in the challenge were lower than those reported in other studies. This was anticipated, considering the nature of the specific modality---respiration. Other modalities, such as behavioral (\textit{e.g}., facial videos), physiological (\textit{e.g}., electrodermal activity), or brain activity (\textit{e.g}., fNIRS), demonstrated higher performance.
Regardless, we believe that respiration is an important and underexplored modality, particularly in the context of automatic pain assessment. Furthermore, we emphasize that respiration is a strong candidate for remote, contactless patient monitoring. Other modalities, such as facial videos or pseudo-cardiac signals derived from them, are sensitive to common challenges in clinical environments, including facial occlusions or temporary disappearance from view. In contrast, respiration can be captured using vision or radar sensors, independent of conditions such as lighting, occlusions, or bed coverings \cite{hari_kumar_2025, yang_liu_2025}.
We suggest that future research should explore the use of respiration signals for automatic pain assessment, either as a standalone modality or in combination with other modalities.

\section*{Safe and Responsible Innovation Statement}
This work relied on the \textit{AI4PAIN} dataset \cite{rojas_hirachan_2023,ai4pain_2024,ai4pain_2025}, made available by the challenge organizers, to assess automatic pain recognition methods. All participants confirmed the absence of neurological or psychiatric conditions, unstable health issues, chronic pain, or regular medication use during the session. Before the experiment, participants were thoroughly informed of the procedures, and written consent was obtained. The original study's human-subject protocol received ethical clearance from the University of Canberra's Human Ethics Committee \textit{(approval number: 11837)}. 
The proposed method was aimed at continuous pain monitoring, though its clinical use demands validation beyond controlled settings.


\section*{Acknowledgements}
This paper is supported by the projects that have received funding from the
European Union's Horizon 2020 research and innovation programme under grant agreement
$101080905$ (\textit{STRATIFYHF project}).

\bibliographystyle{ACM-Reference-Format}
\bibliography{library}

\end{document}